\documentclass[conference]{IEEEtran}
\usepackage{cite}
\usepackage{amsmath,amssymb,amsfonts}
\usepackage{algorithmic}
\usepackage{graphicx}
\usepackage{textcomp}
\usepackage{multicol}
\usepackage{multirow}
\usepackage{cleveref}
\usepackage{censor}
\usepackage{tablefootnote}
\usepackage{lipsum} 
\usepackage{xcolor}
\usepackage{url}
\usepackage{soul}
\def\BibTeX{{\rm B\kern-.05em{\sc i\kern-.025em b}\kern-.08em
    T\kern-.1667em\lower.7ex\hbox{E}\kern-.125emX}}

\begin{document}

\title{Towards Characterizing Cyber Networks with Large Language Models
\thanks{}
}
\author{\IEEEauthorblockN{Alaric Hartsock}
\IEEEauthorblockA{\textit{National Security Directorate} \\
\textit{Pacific Northwest National Laboratory}\\
Richland, WA, USA \\
alaric.hartsock@pnnl.gov}
\and
\IEEEauthorblockN{Luiz Manella Pereira}
\IEEEauthorblockA{\textit{Global Security Directorate} \\
\textit{Savannah River National Laboratory}\\
Aiken, SC, USA \\
luiz.pereira@srnl.doe.gov}
\and
\IEEEauthorblockN{Glenn Fink}
\IEEEauthorblockA{\textit{Global Security Directorate} \\
\textit{Savannah River National Laboratory}\\
Aiken, SC, USA \\
glenn.fink@srnl.doe.gov}
}
\maketitle

\begin{abstract}

Threat hunting analyzes large, noisy, high-dimensional data to find sparse adversarial behavior. We believe adversarial activities, however they are disguised, are extremely difficult to completely obscure in high dimensional space. In this paper, we employ these latent features of cyber data to find anomalies via a prototype tool called Cyber Log Embeddings Model (CLEM). CLEM was trained on Zeek network traffic logs from both a real-world production network and an from Internet of Things (IoT) cybersecurity testbed. The model is deliberately overtrained on a sliding window of data to characterize each window closely. We use the Adjusted Rand Index (ARI) to comparing the k-means clustering of CLEM output to expert labeling of the embeddings. Our approach demonstrates that there is promise in using natural language modeling to understand cyber data.
\end{abstract}


\section{Introduction}
Threat hunting is an open-ended cybersecurity exploration to detect abnormal behaviors that automated tools cannot easily find. Threat hunters continuously collect and analyze data from their corporate networks looking for indicators of compromise. The complexity and changing attack surface of industrial control systems creates an increasing need for tools to help hunters effectively detect threats within diverse, complex, and statistically noisy cyber data. 

This project addresses this gap by developing CLEM, which is capable of ingesting cybersecurity logs from cyberphysical systems and characterizing entities in them by creating vector identities and behavioral relationships. The key innovation is the use of large language models (LLMs) to model the linguistic structures found in cybersecurity logs allowing us to cluster machines according to their behaviors over time.

The rest of this paper is divided as follows: Section \ref{motivations} provides motivation for our work and methodologies. In Section \ref{related_works} we provide a series of related papers spanning different models, approaches, and techniques. We introduce our main contribution, CLEM, in Section \ref{clem} and our results in Section \ref{results}. Finally, we provide future research ideas in Section \ref{future_work} and conclude our work in Section \ref{conclusion}.

\section{Motivations}
\label{motivations}

Threat hunters need to be able to understand how machines, names, protocols, users, and other entities change roles and relationships over time. When machines begin acting abnormally, highlighting these changes can help defenders know where to look.

When applying natural language processing (NLP) to cyber data, it is typical to embed the data into dense vector representations that encode semantic meaning.  Mikolov, et al. \cite{mikolov-etal-2013-linguistic} showed the power of word embeddings to enable algebraic manipulation of word meanings. For example, define $\overrightarrow{x}$ as the embedding vector of the word $x$, we can demonstrate an approximate relationship between \textit{king} and \textit{queen} through a vector difference $\overrightarrow{woman} - \overrightarrow{man}$:
\begin{equation}
    \overrightarrow{king} - \overrightarrow{man} + \overrightarrow{woman} \approx \overrightarrow{queen}, \label{eqn:1}
\end{equation}
Embeddings allow us to vectorize abstract cyber data, such as behaviors of machines, protocols, and connections; their vector representations can be subsequently used to train other neural networks (e.g. \cite{vaswani2023attention}) to classify behaviors of systems as a whole.

Suppose threat hunters are looking for Trojan horse programs in a network. Normally, machine \text{1.2.3.4} runs a mail server program that communicates using port 25. In the Zeek logs this combination would be expressed as "\textit{id.resp\_h: 1.2.3.4, id.resp\_p: 25}" The Cerberus banking trojan malware \cite{mitre_anchor} communicates to its remote server using Transmission Control Protocol client port 8888. If 1.2.3.4 has been compromised by this malware, we might see "\textit{id.orig\_h: 1.2.3.4, id.orig\_p: 8888}". Although port 8888 is often used by web protocols this behavior would be different from normal for this machine. When this behavior appears, the embeddings for the infected machine will separate from the normal cluster of email servers. Similarly to equation \ref{eqn:1}, we can leverage vector arithmetic on the embeddings to study these relationships.



Let a service running on machine $M$ normally communicate using port, $p$. We represent the embedding vector of this combination as $\overrightarrow{M_p}$. Now let $q$ be an unusual port possibly associated with malware. Thus, $M_p$ would be the normal behavior and $M_q$ would be the malicious behavior. Even if $M$ is not known to be infected, armed with equation \ref{eqn:1}, we may be able to determine what it, and any other unknown infected machine, would look like if it were:
\begin{align}
    \overrightarrow{M_p} - \overrightarrow{p} + \overrightarrow{q}\approx \overrightarrow{M_q} \label{eqn:2}
\end{align}



\section{Related Works}
\label{related_works}

There have been many applications of machine learning (ML) and natural language processing to cybersecurity. With recent advancements in NLP, researchers aiming at improving, or automating, existing pipelines, have applied these novel techniques in areas such as threat intelligence, phishing detection, malware detection, log analysis, and more. Various methodologies rely on expert-engineered features that are used by ML models to perform classification. LLMs have enabled entire logs to be vectorized, creating feature vectors that are more information-dense and allowing these temporal models to yield state-of-the-art results.

In \cite{Koda2020Anomalous}, Koda et al. used IP2Vec and Support Vector Data Description (SVDD) to aid security operations center operators to identify malicious or compromised Internet Protocol (IP) addresses in large volumes of network logs. Their model was trained on flow-based traffic logs from the External Server week2-week4 datasets of Coburg Intrusion Detection Data Sets (CIDDS)-001 \cite{ring2017flow}; it contains network traffic captured in an emulated small business environment including normal activity, port scans, brute-force password-guessing attacks, and alert logs from intrusion detection systems. By using SVDD to tune the IP2Vec features, the authors used IP address embeddings to detect the attackers hidden within the logs.

Contrastively, Hammerschmidt et al. presented a method for behavioral clustering of IP flow record data to identify different activities and behaviors in network traffic \cite{Hammerschmidt2016Behavioral}. By detecting changes in communication behavior, their model builds accurate profiles by detecting concept drift. The authors introduce the concept of freshness which is used to determine when a host changes its behavior by identifying change-points. Paired with probabilistic deterministic finite automata, the authors are able to categorize IP flow records accurately for both synthetic and real botnet data.

More recent works focus on the power of word embeddings as a driving technology. In \cite{Asudani2023}, the authors perform a survey-styled comparison between embeddings and other analytics and demonstrate the superiority of embeddings on text-based cyber data. Asudani et al. concluded embeddings provide various benefits, including the ability to: learn latent relationships, produce dense representations (compared to sparse methods like bag-of-words), enable algebraic analytics, scale to large datasets, and bypass the need for expert-engineered features.

LLMs have opened a new avenue of research for cyber researchers. Not only can LLMs process logs entirely, but they also train state-of-the-art embedding mechanisms. There are two use cases for LLMs in processing cyber data: either the model is used for classification or the model is fine-tuned and only used to generate embeddings.

In \cite{alkhatib2022can}, Alkhatib et al., train a Bidirectional Encoder Representations from Transformers (BERT) \cite{devlin2018bert} in a self-supervised masked fashion to reconstruct Controller Area Network (CAN) Identifiers (CAN IDs), which are data frames containing diagnostic, informative, and controlling data that have been encoded into an identifier. Trained using a masking technique, at inference, the authors mask parts of the incoming data, use the fine-tuned BERT model to predict the masked data, and determine if the original input is anomalous by checking if the predicted counterpart is among anticipated candidates. On the other hand, Manocchio et al., introduce FlowTransformer, \cite{manocchio2024flowtransformer}, a pipeline that facilitates swapping different parts of the LLM framework to expedite the creation, training, and testing of new models for network intrusion detection systems. The authors' workflow constitutes fine-tuning a large language model on network data and using it to predict anomalous flow data. While they tested various different models, two of the most notable are Generative Pre-Trained (GPT) 2.0 and BERT. In a similar fashion, Karlsen et al, perform a benchmark comparison between BERT, Robustly optimized BERT approach (RoBERTa), distilled RoBERTa (DistilRoBERTa), GPT-2, and GPT-Neo to test the models' ability to analyze logs and determine anomalous behavior \cite{karlsen2024benchmarking}. For the datasets Apache Access Dataset, Consejo Superior de Investigaciones Científicas (CSIC) 2010, Practice of Knowledge Discovery in Databases (PKDD), Thunderbird, OpenStack, and Spirit, each model is fine-tuned and used to analyze new logs. The authors show the pre-trained models can leverage the newly learned patterns to identify anomalies. Moreover, the authors provide visual insight by passing embedded logs through t-distributed Stochastic Neighbor Embedding (t-SNE) and SHapley Additive exPlanations (SHAP) visualization methods, which act as explanatory graphics as to why the model categorized logs as normal or anomalous.


In contrast with using language models to make predictions, the following papers utilize LLMs for the embeddings they generate. These embeddings are dense vector representations that can be utilized as a feature for downstream pipelines. For example, in \cite{gniewkowski2023sec2vec}, the authors compare using Bag-of-Words (BoW), fastText, and RoBERTa embeddings as features used to train a classifier. Two classification models are trained, one to detect anomalies in Hypertext Transfer Protocol (HTTP) requests and the other to detect malicious uniform resource locators. Similarly, Montes et al., use RoBERTa to embed HTTP requests and use them to train a neural network that predicts if the headers are potential attacks and compare their results to a classic rule-based ModSecurity firewall configured with the Open Worldwide Application Security Project Core Ruleset \cite{montes2021web}. Lastly, \cite{wang2018anomaly} and \cite{seyyar2022detection} both introduce the same pipeline architecture with few differences. While the former uses Word2Vec to obtain embeddings and train a Long Short-Term Memory (LSTM) layer, the latter uses BERT and trains a Convolutional Neural Network.

The current use of language models in cybersecurity has relied on supervised techniques. While in this work we leverage LLMs to embed cyber data, we introduce an unsupervised step that clusters data with similar behaviors. Visualizing these clusters allows us to show behavioral changes of machines in a network over time. 

\section{CLEM}
\label{clem}

CLEM is a threat hunting analysis tool that uses BERT to derive embeddings from Zeek connection (conn) logs  to automatically classify entities in a computer network by their behavior. CLEM's embeddings store behavioral semantics of a computer network at a particular time in high-dimensional space. By dimensionally-reducing and plotting these embeddings, we find that they cluster meaningfully in arrangements that can provide threat hunters with additional information about devices in a network.

\subsection{Data Collection and Processing}

CLEM has been trained on several data sources including the two we discuss here: (1) virtual private network (VPN) from Pacific Northwest National Laboratory (PNNL) including internal and external traffic that passed through the PNNL VPN gateway and (2) the Army Cyber Institute (ACI) IoT Network Traffic Dataset \cite{bastian2023aci}, an open-source dataset with simulated attacks on a small network of IoT devices. Both datasets were originally full packet capture binary data, which we processed into text as Zeek flow logs. When ingesting the training data, we omit fields such as unique flow identifiers and time stamps (\textit{UID} and \textit{ts} fields) because these produced too many meaningless tokens, or created random strings of numbers that diluted the meanings of important integers like IP address octets or port numbers.

\textit{PNNL dataset}: When the pandemic hit in 2020, over 5,000 users began accessing the laboratory exclusively via the VPN and could no longer be characterized by their location on campus. CLEM was commissioned as one way to understand what should be normal in this data set.

\textit{ACI dataset}: The primary objective for ACI was to create a realistic dataset tailored for ML applications IoT network security. Realism of the data aside, this data is simulated. The activities of the devices are not driven by any human actors, and thus the kinds of behaviors that can be found are necessarily limited.

CLEM is designed to find long-term behavioral changes in streaming network data. We simulate streaming by dividing the data into overlapping time windows of a given duration or number of connections. Then, we train over each window in batches until a desired degree of loss is achieved to characterize normal relationships over that time period. We subsequently display plots of the windows in order to show the changes in relationships over time. We refer to this as our Network Storm Tracker, which will be expanded upon in future publications.

\subsection{Design} 

We used a standard BERT (bert-base-cased) model starting from random weights. Because tokenization using human-expert-defined vocabulary rules grows without bounds (there are $2^{32}$ possible IPv4 addresses alone), we use the WordPiece tokenization originally described by Wu, et al. \cite{wu2016googles}; this reduced the token dictionary to a constant 30,522 tokens. We found that the standard BERT tokenizer was very inefficient for Zeek data, creating many more tokens per line than necessary so we therefore decided to train a new tokenizer on a set of Zeek data.

We begin training by reading a window of $10,000$ lines of streaming data and training on it for multiple epochs until it achieves a cross-entropy loss $<0.02$. Then we advance the stream by $5,000$ log lines and train again. While training on the first window takes the longest, subsequent windows require, on average, only a single epoch to achieve the target threshold. Once we have covered the entire stream, training stops.

Overfitting is a common problem where a model approximates the training data too closely, but poorly on unseen data. While it is typically avoided when training a model, we deliberately designed CLEM to overfit each window because it allows us to tightly adjust CLEM's embeddings to characterize the window of data that is being analyzed.

\label{generating-embeddings}
Once our model is trained, we can extract embeddings from it by running text through the model. To get the embedding for a connection, we embed its 5-tuple subset as it would appear in Zeek format, containing the following fields: Originating IP, Originating Port, Responding IP, Responding Port, Protocol.



After embedding the 5-tuples and IP addresses, both are reduced to 2 or 3 dimensions using Uniform Manifold Approximation (UMAP) \cite{mcinnes2020umapuniformmanifoldapproximation}. Based on our experience with these models and data, we believe that the apparent spatial clustering of the address or connection entities relates to their behavior in the network. If true, movement of an entity between clusters is of special significance to threat hunters, meaning that the entity changed behavior. We color the embeddings according to expert assessment of the class of the IP address, usually by subnet (PNNL data) or known hardware type (ACI data).


\section{Results}

While CLEM generates embeddings that we dimensionally reduce, we need to assign labels to the lower-dimensional points. We accomplish this goal by clustering with K-means, which assigns labels based on proximity. Then, we can use expert-derived labels as a second clustering of the data and compare them using the ARI \cite{Hubert1985}, which measures the degree of similarity between two clusters of data while adjusting for the random grouping of elements. Whereas Rand Index (RI) is given by {$\text{RI} = (a + b)/\binom{n}{2}$}, where \textit{a} is the number of pairs of elements that are in the same cluster in both clusterings, \textit{b} is the number of pairs of elements that are in different clusters in both clusterings, and \(\binom{n}{2}\) is the total number of pairs of elements, ARI is defined as:

\begin{equation}
    \text{ARI} = \frac{\text{RI} - \mathbb{E}[\text{RI}]}{\max(\text{RI}) - \mathbb{E}[\text{RI}]}
    \label{ari_equation}
\end{equation}

A negative ARI indicates dissimilar cluster memberships, while a positive ARI implies some similarity between clusters. An ARI score of exactly zero is expected if one of the clusterings is random. An ARI value of 1 indicates identical cluster memberships.


To determine the number of clusters for our K-means clustering approach, we selected the value which produced the highest ARI where $1 <= k <= N$ and $N$ is the number of expert-derived clusters. For example, in Fig.\ref{fig2}, we witness an ARI peaking at  \(\approx 0.82 \) with 4 clusters.


\label{results}

\begin{figure}[tbp]
\includegraphics[width=\columnwidth]{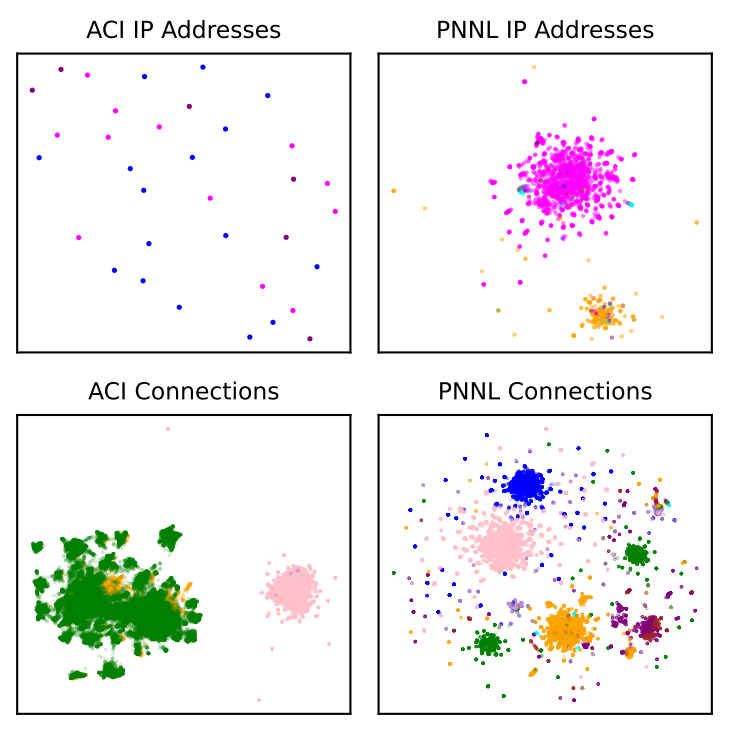}
    \caption{
    Dimensionally reduced connection and address embeddings for the PNNL and ACI data. Colors are assigned to groups that appear in the data and there are many more categories than distinguishable colors. We color the nodes to show the degree of homogeneity of the clusters only.}
\label{fig1}\
\end{figure}

\begin{figure}[htbp]
\centerline{\includegraphics[width=\linewidth]{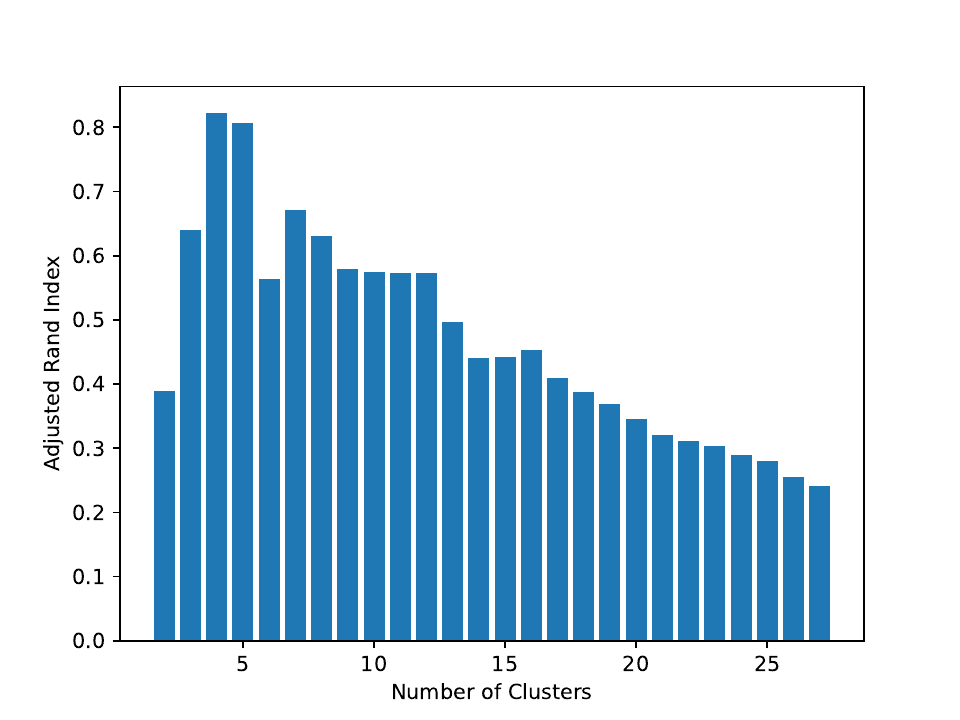}}
\caption{Calculated Adjusted Rand Index vs Number of Clusters}
\label{fig2}
\end{figure}

Embedding-based clustering of cyber data from the PNNL VPN reveals distinct groupings that significantly correlate with expert-annotated machine and connection types. Table \ref{results_table}, visualized in Fig.\ref{fig1} demonstrates a strong agreement between the optimal $k$ clusters and expert labels, achieving a max score of 0.82. Further investigation revealed that these clusters align with known categories such as server types, workstation roles, and communication protocols.

We note that total agreement with the expert labels would mean that our embedding approach had merely discovered what the experts already knew. Notably, the PNNL data is from a production network with real users and labels that experts use to capture behavior. The ACI data is real in the sense that real devices were used, but it was artificial in that no human activity was captured. The ACI labels are only of device type, not intended or actual use. Even the attacks were only automated tools performing denial of service attacks according to a tool-determined scheme. As we would expect then, the ARI showed much more randomness for the ACI IP address data. We believe embeddings capture intentional behaviors, but we have not yet been able to prove this conclusively.

\begin{table}
\centering
\caption{ARI Cluster Comparison Between Datasets}
\begin{tabular}{|c|c|c|c|c|c|}
\hline
 & \textbf{Type of} & \textbf{\# of} & \textbf{K-means} & \textbf{Expert} & \\
\textbf{Dataset} & \textbf{Embedding}  & \textbf{Embeddings} & \textbf{Clusters} &\textbf{labels} & \textbf{ARI} \\ \hline
\multirow{2}{*}{PNNL}&IP Addresses & 1900 & 2 & 24 & 0.68 \\ \cline{2-6}
&Connections  & 100,000 \tablefootnote{\label{shared} In total there were a few million connection embeddings in either dataset. We extracted a random sample of 100,000 of them to reduce analytical complexity} & 4 & - \tablefootnote{\label{expert}There are a very large number of possible expert labels for connections, and we did not count them.} & 0.82  \\ \hline
\multirow{2}{*}{ACI}&IP Addresses & 42 & 6 & 4 & 0.06 \\ \cline{2-6} &Connections & 100,000 \footref{shared} & 2 & - \footref{expert} & 0.65 \\ \hline
\end{tabular}
\label{results_table}
\end{table}


\section{Future Work}
\label{future_work}
In this section we outline several areas of likely future development of the CLEM tool and its applications.  The windows of embeddings in a stream could be displayed as an animation of the changing network conditions over time. We call this our Network Storm Tracker after our desire to emulate a weather display for threat hunters. Moreover, aligned UMAP can be used to normalize embedding movement between embedding plots. However, it creates conditional placements of each marker on all the others, and this causes entire clusters to continually move across the screen. When scores of clusters move apparently randomly it becomes difficult to interpret behavior from relative position. We need to devise a method of movement normalization that allows what is normal movement to be distinguished from abnormal. For instance, movement of a marker between clusters is probably indicative of behavior change while coordinated movement of the entire cluster of machines or connections is probably meaningless.

To derive the true potential of embeddings to capture semantic meaning of cyber data, we plan to collect more data from real systems where ground-truth activity is known. This will allow us to contrast the effects of intentional actions in a real network from stochastic events in a simulated environment.
    

\section{Conclusion}
\label{conclusion}

We have outlined a novel approach to embedding cyber logs with the potential to represent the semantic relationships of cyber entities as the latent relationships between their embeddings. The result will be a characterization of true behavior that could greatly improve the abilities of threat hunters to find their elusive adversaries. Just as human bias has been detected in LLM behavior, adversarial ``bias'' may be difficult to hide in the high dimensional meanings captured by text embeddings of cyber data.

Our technique is novel in that we embed and visualize cyber terms for the sole purpose of human threat hunter analysis instead of for use in a downstream model. We train a BERT model on a stream of Zeek conn logs, deliberately overfitting in order to learn the underlying structure of the data. Using ARI as a metric of agreement between our embeddings clustered with K-means and expert labelling, we witness meaningful similarity, demonstrating the need for further investigation into the meaning and utility of these embeddings. In conclusion, CLEM represents a significant step forward in analyzing and representing the complex behaviors of computer networks.

\section*{Acknowledgment}
This research was supported in part by the “Resilience through Data-driven Intelligently-Designed Control” Initiative, under the Laboratory Directed Research and Development Program at PNNL. PNNL is a multi-program national laboratory operated for the U.S. Department of Energy by Battelle Memorial Institute under Contract No. DE-AC05-76RL01830.

We acknowledge the work of PNNL scientists Jeremy Teuton, Craig Bakker, Gregg Serene, and Casey O'Leary for work that supported this project.

This research was performed in part by an appointee of the Minority Serving Institutions Internship Program (MSIIP) administered by the Oak Ridge Institute for Science and Education (ORISE) for the National Nuclear Security Administration (NNSA) and U.S. Department of Energy (DOE). ORISE is managed by Oak Ridge Associated Universities (ORAU). All opinions expressed herein are the authors' alone, not the positions of of the sponsors.







\bibliographystyle{IEEEtran}
\bibliography{conference_101719}

\end{document}